\documentclass[letterpaper, 10 pt, conference]{ieeeconf}  

\IEEEoverridecommandlockouts                              

\usepackage{booktabs}
\usepackage{graphicx}
\usepackage{multirow}
\usepackage{xcolor}
\usepackage{amssymb}
\usepackage{hyperref}

\title{\LARGE \bf
Indoor Future Person Localization from an Egocentric Wearable Camera
}

\author{Jianing Qiu$^{1,2}$, Frank P.-W. Lo$^{2}$, Xiao Gu$^{1,2}$, Yingnan Sun$^{2}$, Shuo Jiang$^{3}$ and Benny Lo$^{2}$

\thanks{$^{1}$Department of Computing, Imperial College London
        {\tt\small \{jianing.qiu17, xiao.gu17\}@imperial.ac.uk}}%
\thanks{$^{2}$The Hamlyn Centre, Imperial College London
        {\tt\small \{po.lo15, y.sun16, benny.lo\}@imperial.ac.uk}}%
\thanks{$^{3}$College of Electronics and Information Engineering, Tongji University
        {\tt\small jiangshuo@tongji.edu.cn}}%
}

\begin{document}

\maketitle
\thispagestyle{empty}
\pagestyle{empty}

\begin{abstract}

Accurate prediction of future person location and movement trajectory from an egocentric wearable camera can benefit a wide range of applications, such as assisting visually impaired people in navigation, and the development of mobility assistance for people with disability. In this work, a new egocentric dataset was constructed using a wearable camera, with 8,250 short clips of a targeted person either walking 1) toward, 2) away, or 3) across the camera wearer in indoor environments, or 4) staying still in the scene, and 13,817 person bounding boxes were manually labelled. Apart from the bounding boxes, the dataset also contains the estimated pose of the targeted person as well as the IMU signal of the wearable camera at each time point. An LSTM-based encoder-decoder framework was designed to predict the future location and movement trajectory of the targeted person in this egocentric setting. Extensive experiments have been conducted on the new dataset, and have shown that the proposed method is able to reliably and better predict future person location and trajectory in egocentric videos captured by the wearable camera compared to three baselines. Our dataset can be downloaded from \href{https://1drv.ms/u/s!AqPeVPaZH5TTcKv1ZLGT5D634ZA?e=FYHIx9}{\textcolor{blue}{here}}

\end{abstract}

\section{Introduction}

With the recent advances in wearable technologies, wearable cameras are becoming popular and able to capture egocentric (first-person) videos with less motion blur and wider field of view. Wearable cameras have been applied to many human-centric applications, such as gaze prediction to understand human intention and attention~\cite{Li_2018_ECCV,huang2018predicting}, human action recognition~\cite{damen2018scaling,kazakos2019epic,sudhakaran2019lsta}, and also learning environment affordances~\cite{nagarajan2020ego}, in egocentric settings. Accurate prediction of future person location and movement trajectory in egocentric videos captured by a wearable camera can help visually impaired people safely navigate, and can also facilitate the development of accurate and personalized mobility assistance (such as autonomous wheelchairs or exoskeleton robots).

Apart from visual images, some wearable cameras are also able to record other sensing signals, such as audio, GPS and IMU, synchronously. Fusing these signals recorded by the wearable cameras could enable better navigation assistance for the user, and may also enable the transfer of assistance developed using wearable cameras to robots. 
For a social or service robot deployed in an indoor environment, for instance, in a large shopping mall, accurate prediction of the future locations and movement trajectories of people in the vicinity can help the robot better plan its navigating path, avoid potential collisions with people, and if needed, track and follow a targeted person more precisely.

Most works on future person location and trajectory prediction, however, have only been focused on scenes captured by birds-eye or oblique views~\cite{pellegrini2009you,kitani2012activity,huang2016deep,yi2016pedestrian,alahi2016social,robicquet2016learning,ma2017forecasting,gupta2018social,liang2019peeking,choi2019looking,liang2020garden}, or from vehicle-mounted cameras~\cite{kooij2014context,bhattacharyya2018long,styles2019forecasting,chandra2019traphic,makansi2020multimodal,mangalam2020disentangling}. Little research has been carried out in egocentric scenarios captured by wearable cameras~\cite{soo2016egocentric,Yagi_2018_CVPR,huynh2020aol}. In the use of a wearable camera, the motion of the camera wearer plays an important role in predicting the future location and movement trajectory of a targeted person relative to the wearer's viewpoint, which is very different from scenarios captured by fixed cameras. Although both wearable and vehicle-mounted cameras provide egocentric vision, vehicles are mainly on the road, whereas a camera wearer can be in many different indoor and outdoor places, and the wearer can change directions more flexibly and abruptly compared to a vehicle. Furthermore, the use of deduced information from vehicle-mounted and wearable cameras is different. One is for assisted or autonomous driving, and the other can be used for human mobility assistance. Egocentric vision from a wearable camera reflects how the camera wearer perceives their surroundings, and with synchronized IMU signals, the intent of the camera wearer can be better inferred. For example, whether the wearer is going to change his/her path or to stop to avoid collision. Therefore, the prediction of future location of a targeted person relative to the wearer's viewpoint can be more accurate. Egocentric vision also provides a close up view of the people in the camera wearer's vicinity, which offers additional cues for predicting their future locations, such as the targeted person's pose~\cite{Yagi_2018_CVPR,mangalam2020disentangling}, and person-person interactions. In this work, we aim to predict the future location and trajectory of a targeted person in videos captured by a wearable camera in indoor environments.

The contributions of this work are twofold:
\begin{itemize}
  \item A novel egocentric video dataset was constructed for future person location and movement trajectory prediction. The dataset contains over 8k short clips. For each targeted person, a tight bounding box was manually labelled in each frame, resulting in nearly 14k bounding boxes in total. The estimated pose of each targeted person in each frame is also included along with the synchronized 6-axis IMU signals of the wearable camera. Although this dataset was constructed for future location and trajectory prediction, it can also be used to develop and evaluate a tracking system in the egocentric setting, and it also complements the existing FPL ~\cite{Yagi_2018_CVPR} and Citywalks~\cite{styles2020multiple} datasets, whose annotations are based on automated detection methods. To the best of our knowledge, this is the first egocentric dataset that provides human annotated person bounding boxes, detected human body poses, and synchronized IMU signals. The dataset is applicable to multiple tasks and will be made available upon request.
  \item An LSTM-based encoder-decoder framework was designed for the prediction of the location and movement trajectory of a targeted person, which takes the previous locations and poses of the targeted person, and IMU signals of the camera as the input, and outputs the predicted future location and movement trajectory of the targeted person.
\end{itemize}

\section{Related Work}

Future person location and movement trajectory prediction have been an active research area, but predictions in most studies have only been carried out in videos captured from birds-eye or oblique views~\cite{pellegrini2009you,kitani2012activity,huang2016deep,yi2016pedestrian,alahi2016social,robicquet2016learning,ma2017forecasting,gupta2018social,liang2019peeking,choi2019looking,liang2020garden}. Social-LSTM was proposed in~\cite{alahi2016social} to predict human trajectory, which uses social pooling layers to model the effect of nearby people's behaviours on a person's future trajectory. Social-GAN~\cite{gupta2018social} further introduces adversarial learning in future trajectory prediction. In~\cite{liang2019peeking}, a person's future trajectory is jointly predicted with his/her future activities using rich visual semantic features, such as scene and objects. Predicting multiple possible future trajectories of people have also been studied in~\cite{liang2020garden}. However, these approaches are not targeted for egocentric scenarios, in which the motion of the camera plays an important role, and the camera view is also largely different from the top-down or oblique views. In the following, we review two egocentric scenarios relevant to or involving future person localization.

\textbf{Vehicle-mounted cameras.} With the use of visual data captured by vehicle-mounted cameras, predicting pedestrian trajectory has been an active research area in autonomous driving~\cite{kooij2014context,bhattacharyya2018long,styles2019forecasting,chandra2019traphic,makansi2020multimodal}. Studies have also been carried out in predicting pedestrian's intent of crossing or not crossing the street~\cite{gujjar2019classifying,liu2020spatiotemporal}, forecasting pedestrian locomotion~\cite{mangalam2020disentangling}, and detecting traffic accident with pedestrian future location prediction~\cite{iros2019yao}. Although not targeted at pedestrians, \cite{icra2019yao} and \cite{malla2019nemo} have studied future vehicle localization, as their works are targeted at vehicles, the designed framework and the used cues for future localization differ from ours (e.g., we are able to use human pose as a cue, similar to those in~\cite{Yagi_2018_CVPR,mangalam2020disentangling}). All works above are part of an assisted or autonomous driving system, whereas our work focuses on using wearable cameras to provide future person localization. 

\textbf{Wearable cameras.} Predicting the future location and trajectory of the camera wearer in egocentric stereo images have been studied in~\cite{soo2016egocentric}. Our work is inspired by~\cite{Yagi_2018_CVPR}, which is the first work that predicts future locations of people (not the camera wearer) in egocentric videos from wearable cameras. However, our work differs from them in that 1) we aim to predict a bounding box as the future location of the targeted person instead of predicting a point as the location as in their work. A bounding box can better reflect how much physical space the person occupies, and therefore can enable better navigation. The predicted bounding box can also facilitate tracking if the method is integrated into a tracking system; 2) IMU signals extracted from the wearable camera are used, along with manually labelled human bounding boxes that can represent the scale of a targeted person; 3) we propose an LSTM-based encoder-decoder framework for the future location and movement trajectory prediction in egocentric videos whereas their predictions are based on a 1D convolution-deconvolution framework (conv-deconv). Built on the conv-deconv framework of~\cite{Yagi_2018_CVPR}, an adaptive online learning scheme is proposed in~\cite{huynh2020aol} to further improve the prediction accuracy. Another relevant work is~\cite{styles2020multiple}, which proposes a GRU-CNN based framework (STED) and the Citywalks dataset. STED encodes past optical flow and bounding boxes of a pedestrian, and then decodes his/her future bounding boxes. Although we both use bounding boxes, we differ from them in the framework for prediction, and we use human pose and IMU data as extra cues for predicting future person location.

\section{Method}

\begin{figure*}[!t]
\centerline{\includegraphics[width=0.9\textwidth]{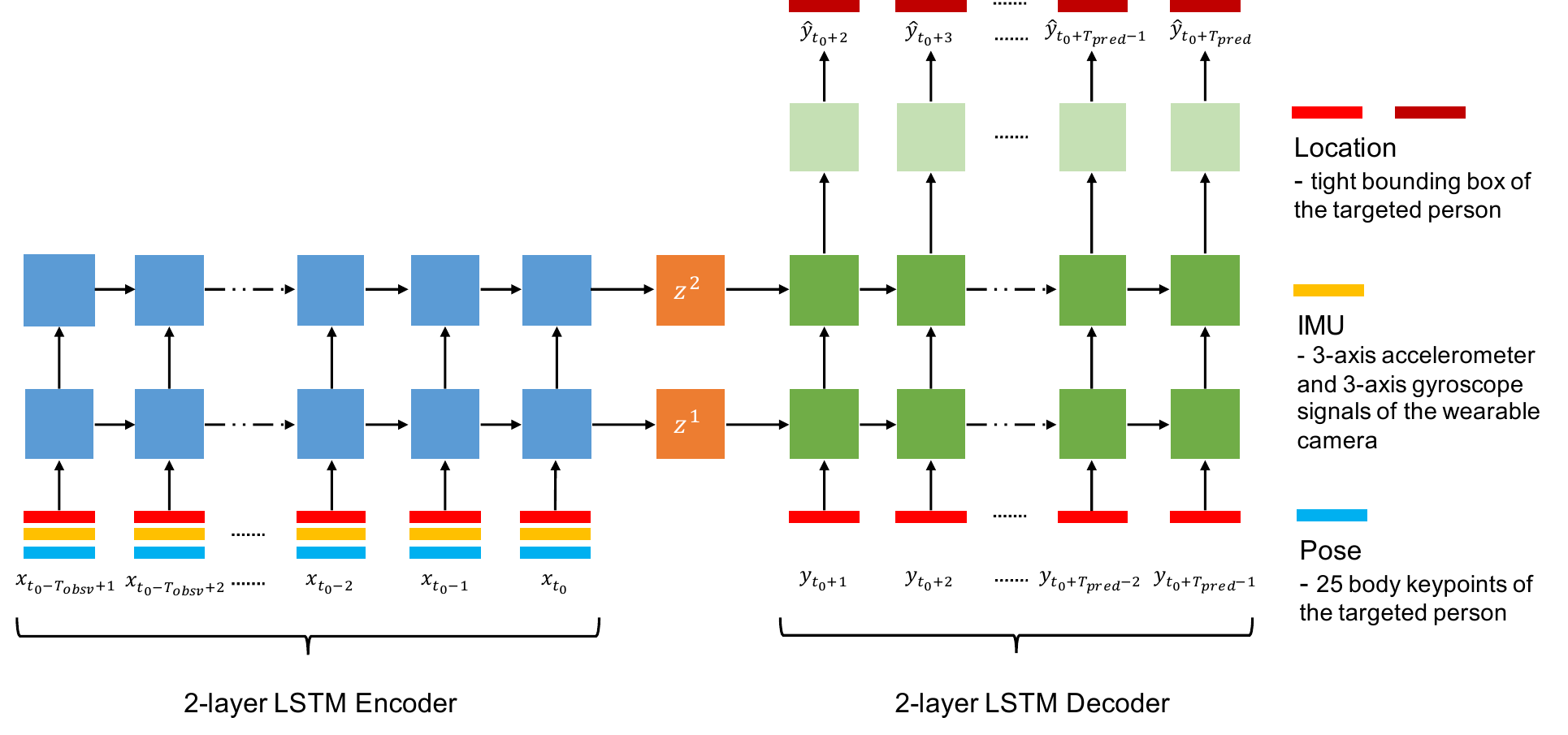}}
\caption{Overview of the LSTM-based encoder-decoder framework for future person location and trajectory prediction. The encoder takes the past locations, body poses of a targeted person, and the past motion of the camera wearer as the input, and encodes them into context vectors $z^{1}$ and $z^{2}$. The decoder receives the context vectors, and decodes the locations of the targeted person at future time steps. We use the bounding box of the targeted person to represent his/her location at each time step, and the person's movement trajectory is formed using the centers of the bounding boxes. Note that in our experiments, the decoder does not receive the true location at $t_{0}$. Therefore, it starts decoding the future location from time step $t_{0}+2$.}
\label{fig:network_architecture}
\end{figure*}

In this work, we aim to predict the locations of a targeted person in future frames of an egocentric video based on the observation of his/her locations, body poses, and the motion of the camera wearer in the past frames. Formally, denoting each frame as a time step $t$, the location of the targeted person in each frame is defined as $l_{t} \in \mathbb{R_{+}}^{1\times4}$ (i.e., the coordinates of the top-left and bottom-right corners of the person bounding box), the body pose in each frame is defined as $p_{t} \in \mathbb{R_{+}}^{K\times2}$ where $K$ is the number of body keypoints, and the motion of the camera wearer in each frame is defined as $i_{t} \in \mathbb{R}^{1\times6}$ (i.e., 6-axis IMU signals from a 3-axis accelerometer and a 3-axis gyroscope). Given an observation period $T_{obsv}$, a prediction period $T_{pred}$, and the time step of the current frame $t = t_{0}$, our goal is to predict a set of future locations $L_{pred} = (l_{t_{0}+1}, ...,  l_{t_{0}+T_{pred}})$ of the targeted person based on his/her past locations $L_{obsv} = (l_{t_{0}-T_{obsv}+1}, ...,  l_{t_{0}})$ and past body poses $P_{obsv} = (p_{t_{0}-T_{obsv}+1}, ...,  p_{t_{0}})$, and the camera wearer's past motion $I_{obsv} = (i_{t_{0}-T_{obsv}+1}, ...,  i_{t_{0}})$. To achieve this, a 2-layer LSTM-based encoder-decoder framework is designed, which is shown in Figure~\ref{fig:network_architecture}. LSTM~\cite{hochreiter1997long} is able to model long-term time evolution, and therefore is suitable for the task of future location and trajectory prediction. The encoder takes $x_{t} = l_{t} + i_{t} + p_{t}, t \in (t_{0}-T_{obsv}+1, ..., t_{0})$ as the input per time step of recurrent calculation. Note that $l_{t}$, $i_{t}$, and $p_{t}$ are first flattened and then concatenated, which makes $x_{t} \in \mathbb{R}^{1\times(10+2\times K)}$. The initial hidden and cell states are both set to 0 (i.e., when $x_{t} = x_{t_{0}-T_{obsv}+1}$).

The final hidden and cell states from both layers of the encoder, denoted as context vectors $z^1$ and $z^2$, are then used as the initial hidden and cell states of the respective layers of the decoder. The decoder then decodes the location of the targeted person in each future frame (i.e., $t \in (t_{0}+1, ..., t_{0}+T_{pred})$). $y_{t} = l_{t}$ is the true location of the targeted person in the time step $t \in (t_{0}, ..., t_{0}+T_{pred}-1)$. The hidden state $s_{t}^{2}, t \in (t_{0}, ..., t_{0}+T_{pred}-1)$ from the top layer of the decoder is fed into a fully connected layer to predict the location $\hat{y}_{t+1}$ at next future time step. The center of the predicted person bounding box is used to form the movement trajectory of the targeted person in the future frames.

We denote the proposed framework as LIP-LSTM as it takes in the location, IMU, and pose as the input.

\section{Dataset}

\begin{table}[]
\centering
\caption{The Number of Samples of Different Walking Directions in the Training and Test Sets}
\label{tab:dataset}
\resizebox{0.9\linewidth}{!}{%
\begin{tabular}{@{}lccccc@{}}
\toprule
Dataset & Toward & Away & Across & Still & Total \\ \midrule
Train & 496    & 5440 & 320    & 388   & 6644  \\
Test  & 253    & 1187 & 65     & 101   & 1606  \\ \bottomrule

\end{tabular}%
}
\end{table}

\begin{figure}[!t]
\centerline{\includegraphics[width=\linewidth]{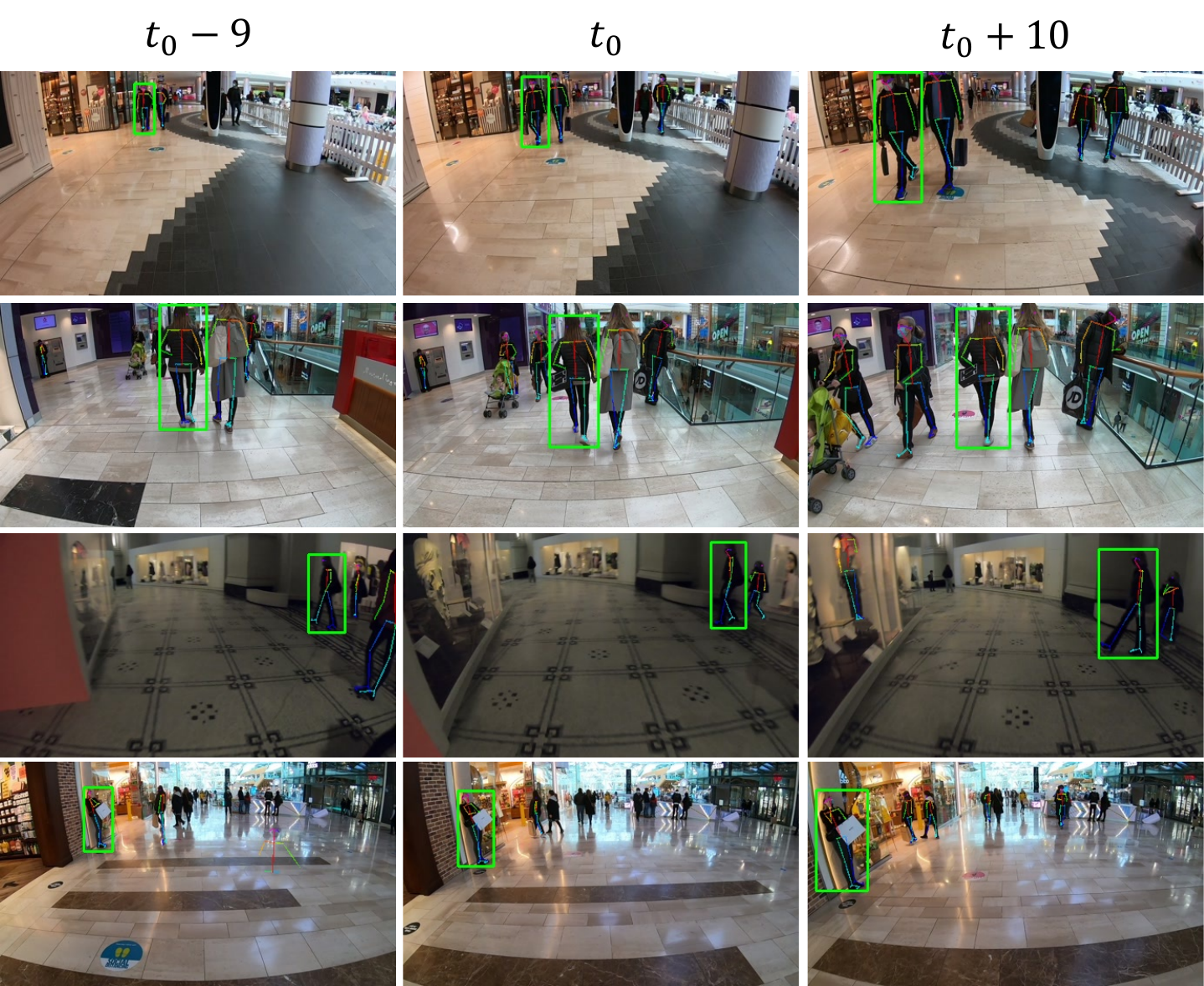}}
\caption{One targeted person is shown for each walking direction. A targeted person at each time step is manually labelled with a tight bounding box and his/her body pose is estimated by OpenPose. From top to bottom, the walking direction of the targeted person with respect to the camera wearer is: 1) \textit{toward}, 2) \textit{away}, 3) \textit{across}, 4) \textit{still}. (Best viewed in color).}
\label{fig:dataset_samples}
\end{figure}

\subsection{Data Collection}

A GoPro Hero 7 Black camera was mounted on the chest of the camera wearer to record egocentric video data in different indoor environments. Data were collected from 2 national museums and 1 large shopping mall while the camera wearer was walking around. All video data were recorded using the wide angle at 30fps with the resolution set to $1920 \times 1080$. 9 videos were recorded from the shopping mall, and 7 videos from the museums. The total length of the recorded data lasts for 1 h 56 m.

\subsection{Data Pre-processing}

The recorded egocentric videos were first downsampled to 10 fps and $455 \times 256$ using ffmpeg\footnote{https://ffmpeg.org/}. OpenPose~\cite{8765346} was then used to detect body poses in each frame after downsampling. We adopted the setting of detecting 25 body keypoints (i.e., $K=25$). The 6-axis IMU (3-axis accelerometer and 3-axis gyroscope) signals associated with each downsampled frame were extracted from GoPro meta data using dashware\footnote{http://www.dashware.net/}. We pre-defined 4 different types of people movements with regards to the camera wearer, among which 3 types are adopted from~\cite{Yagi_2018_CVPR} (i.e., \textit{toward}, \textit{away}, and \textit{across}), and one new type \textit{still} is introduced (i.e., the targeted person is not moving in the scene). We then manually went through the OpenPose processed frames, identifying those 4 types and labelling tight bounding boxes of the targeted people with VIA~\cite{dutta2019vgg}. We label each targeted person for at least 20 consecutive frames (2 seconds at 10 fps), in which this person's body poses are successfully detected by OpenPose. For those longer than 20 frames, a sliding window was applied to generate multiple 2-second samples. In total, there are 8,250 samples obtained, with 13,817 unique bounding boxes. The dataset is then split into training and test sets (training and test samples are drawn from different videos). Table~\ref{tab:dataset} shows the sample distribution across 4 types of walking direction in each set. Figure~\ref{fig:dataset_samples} shows some targeted person samples. We call this new dataset EILT (Egocentric Indoor future Location and Trajectory prediction dataset).

\section{Experiment}

\subsection{Implementation Details}\label{subsec:implementation_details}

We implement our encoder-decoder framework using PyTorch. The hidden dimension of all LSTM layers is 384. Dropout~\cite{srivastava2014dropout} is set to 0.5 for the LSTM layers. The entire network is trained end-to-end with $\ell_2$ loss, which measures the mean squared error between the predicted and true locations. Adam~\cite{kingma2014adam} is used as the optimizer. We set batch size to 64 and train the network for 100 epochs. Teacher forcing is used with a ratio of 0.5 during training. Following~\cite{Yagi_2018_CVPR}, we set both $T_{obsv}$ and $T_{pred}$ to be 10, i.e., a setting of observing 1 second, and predicting for the next second.

\subsection{Baselines}

The following baselines are used to compare the performance of the prediction of a person's location and movement trajectory with our proposed method on the EILT dataset:

\textbf{STATS}: A statistics-based approach. Given a time step $t \in (t_{0}+1, ..., t_{0}+10)$, we first compute the average location $l_{t-1}^{avg}$ of time steps from $t_{0}-9$ up to $t-1$. We then calculate the displacement of the location $l_t$ with respect to $l_{t-1}^{avg}$, which results in a displacement matrix $m \in \mathbb{R}^{10\times4}$. For each type of walking direction (i.e., \textit{toward}, \textit{away}, \textit{across}, or \textit{still}), we average $\sum_{n=1}^{N} m_n$ where $N$ is the total number of samples of a walking direction $j$ to obtain an average displacement matrix $m_{j}^{avg} \in \mathbb{R}^{10\times4}$ of that walking direction. All $m_{j}^{avg}$ are calculated using the training set. During the test, we obtain the walking direction of the targeted person, calculate his/her average location in the past 10 frames, and use the corresponding average displacement matrix of his/her walking direction to iteratively predict his/her locations in the future 10 frames. 
    
\textbf{LR}: We use linear regression to estimate the locations of the targeted person in the future frames. Concretely, linear regression is applied to a set of top-left corners, and a set of bottom-right corners of the person bounding boxes in the observation period, after which two linear functions are obtained, each for predicting two corners of the person bounding boxes in the prediction period. Assuming that the targeted person is moving at a constant speed and does not change his/her walking direction, we compute the average displacement $\alpha$ of the $x$ coordinate of each corner in the observation period. For example, the average displacement of the $x$ coordinate of the top-left corner is calculated as $\alpha^{tl} = (l_{t_{0}}^{tl_x} - l_{t_{0}-9}^{tl_x}) / 9$. The $x$ coordinate of each corner in the prediction period is then incrementally added with respective $\alpha$, starting from the respective $x$ coordinate in the $t_{0}$ frame,  and the $y$ coordinate is estimated using its corresponding linear function.
    
\textbf{L-LSTM}: The same encoder-decoder architecture as the LIP-LSTM with the input to the encoder being only the past locations of the targeted person.

The predictions from $t_{0}+2$ to $t_{0}+10$ of all three baselines are used to compare with our method.

\subsection{Evaluation Metrics}
We adopted the following 3 metrics (also used by~\cite{styles2020multiple}):

\textbf{Mean IOU}: The intersection over union (IOU) averaged across IOUs calculated between the true and predicted bounding boxes of all future time steps starting from $t_{0}+2$.
    
\textbf{Mean Final IOU}: The IOU averaged across those calculated for the final time step (i.e., the final future location at the end of the prediction period $t_{0}+10$).
    
\textbf{Mean DE}: The mean displacement error (DE) is the average $\ell_2$ distance between the true point of a trajectory and the predicted point in each prediction time step~\cite{pellegrini2009you,alahi2016social}. A point of a trajectory is the center of the person's bounding box in our case which starts from $t_{0}+2$.

\subsection{Overall Results}

\begin{figure*}[!t]
\centerline{\includegraphics[width=\textwidth]{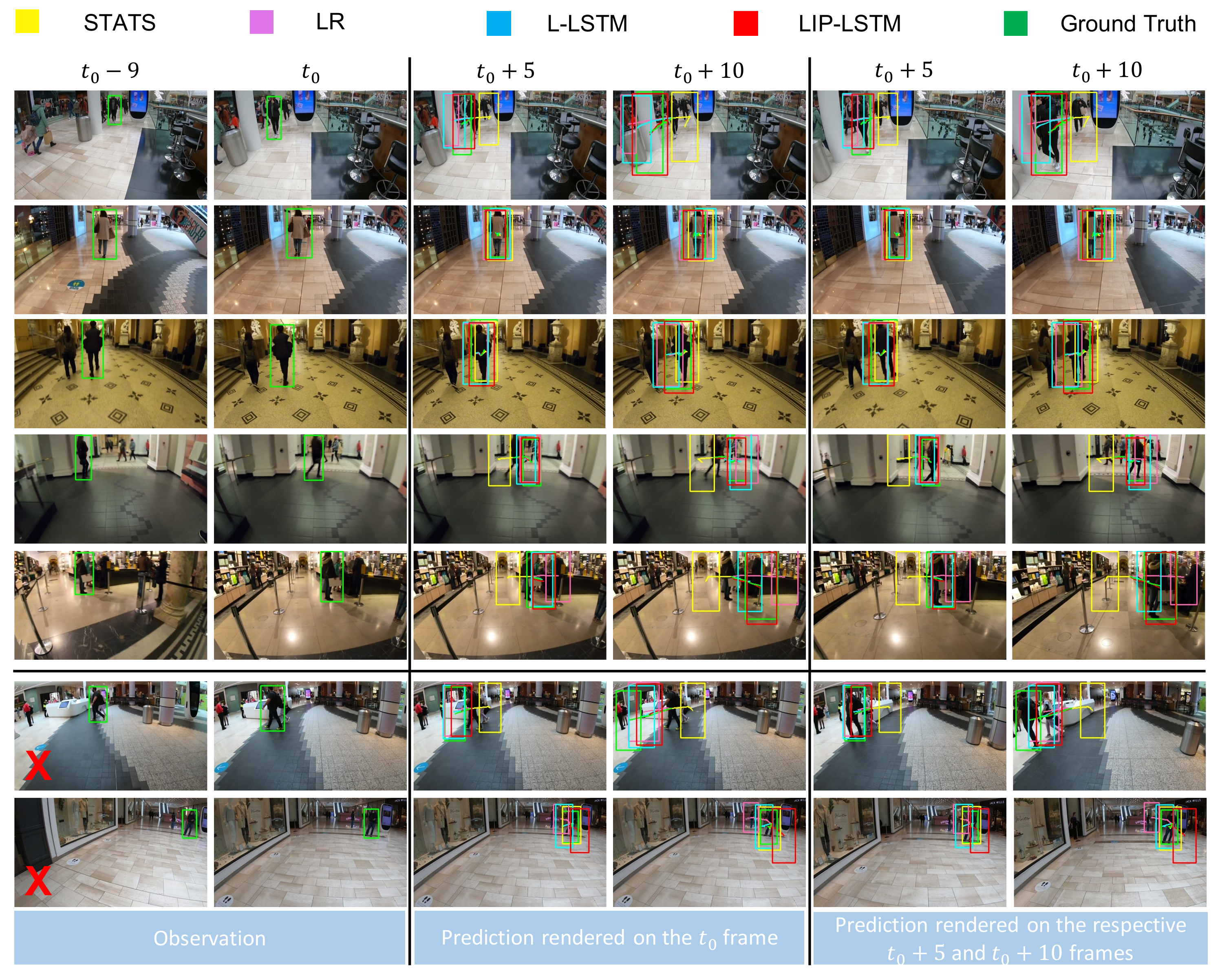}}
\caption{Qualitative results of different methods in predicting future person location and trajectory. Top 5 rows are examples that are successfully predicted by our method (LIP-LSTM). Bottom 2 rows are failure cases. From top to bottom, the walking direction of the targeted person with respect to the camera wearer is 1) toward, 2) away, 3) away, 4) across, 5) still, 6) across, 7) still. Apart from the final location and trajectory at the end of prediction period ($t_{0}+10$), we also show the location and trajectory at the middle of prediction period ($t_{0}+5$).}
\label{fig:vis_train_on_all}
\end{figure*}

Table~\ref{tab:overall_results} shows the overall results and comparisons between the proposed LIP-LSTM and baseline methods. It can be seen that LIP-LSTM shows the best performance in both future location and trajectory predictions as it produces the highest IOUs and the lowest DE. The STATS method has the worst performance in both tasks. With the past locations being the only cue, the LSTM-based encoder-decoder framework is already able to have a 6.6\% relative increase in the measured Mean Final IOU compared to that of LR (0.501 vs 0.470), and reduce Mean DE from 15.8 to 13.9, which is a 12.0\% relative improvement. With the addition of IMU and pose information, the LIP-LSTM is able to further increase the Mean IOU (0.619 to 0.630) and Mean Final IOU (0.501 to 0.513), and reduce the Mean DE from 13.9 to 13.7.

\begin{table}[]
\centering
\caption{Overall Results and Comparisons with Baselines}
\label{tab:overall_results}
\resizebox{0.9\linewidth}{!}{%
\begin{tabular}{@{}lccc@{}}
\toprule
Method        & Mean IOU       & Mean Final IOU & Mean DE        \\ \midrule
STATS         & 0.499          & 0.419          & 24.0          \\
LR           & 0.605          & 0.470          & 15.8          \\
L-LSTM & 0.619          & 0.501          & 13.9          \\ \midrule
\textbf{LIP-LSTM}          & \textbf{0.630} & \textbf{0.513} & \textbf{13.7} \\ \bottomrule
\end{tabular}%
}
\end{table}

\begin{table*}[]
\centering
\caption{Results on Each Walking Direction}
\label{tab:res_each_walking_direction}
\resizebox{\textwidth}{!}{%
\begin{tabular}{@{}l|ccc|ccc|ccc|ccc@{}}
\toprule
\multirow{2}{*}{Method} & \multicolumn{3}{c|}{Toward}                      & \multicolumn{3}{c|}{Away}                        & \multicolumn{3}{c|}{Across}                      & \multicolumn{3}{c}{Still}                        \\ \cmidrule(l){2-13} 
                        & Mean IOU       & Mean Final IOU & Mean DE        & Mean IOU       & Mean Final IOU & Mean DE        & Mean IOU       & Mean Final IOU & Mean DE        & Mean IOU       & Mean Final IOU & Mean DE        \\ \midrule
STATS                   & 0.252          & 0.157          & 41.8         & 0.602          & 0.525          & 15.7          & 0.033          & 0.000          & 64.5         & 0.208          & 0.106          & 50.2         \\
LR                     & 0.439          & 0.223          & 25.7          & 0.665          & 0.551          & 12.5          & 0.477          & 0.310          & 19.0          & 0.403          & 0.245          & 29.0         \\
L-LSTM           & 0.524          & \textbf{0.420} & 19.8          & 0.663          & \textbf{0.555} & 11.8          & 0.496          & \textbf{0.412} & 19.1          & 0.318          & 0.249          & 34.2         \\ \midrule
\textbf{LIP-LSTM}           & \textbf{0.534} & 0.389          & \textbf{18.5} & \textbf{0.667} & 0.551          & \textbf{11.4} & \textbf{0.536} & 0.384          & \textbf{15.9} & \textbf{0.448} & \textbf{0.370} & \textbf{23.8} \\ \bottomrule
\end{tabular}%
}
\end{table*}

Figure~\ref{fig:vis_train_on_all} shows some qualitative results. We visualize the predicted and true locations and trajectories at both the middle ($t_{0}+5$) and end ($t_{0}+10$) of the prediction period. In the first row, in which the targeted person is walking toward the camera wearer, the predicted trajectory of LIP-LSTM well matches the person's true trajectory at both the middle and end time point (please refer to the red and green lines), and its predicted future person locations better align with the ground truth than other three baseline methods'. The predicted trajectories of all three baselines in this case deviate from the true trajectory, and the predicted locations are all behind the ground truth. In the case of the targeted person walking away from the camera wearer, we show 2 examples in the second and third rows. In the second row, the targeted person is walking at the approximately same speed as the camera wearer, and therefore, this person appears in roughly the same location in the future frames. The predicted locations of all methods are close to the ground truth at $t_{0}+5$. However, at $t_{0}+10$, only LIP-LSTM and LR are still able to correctly predict the person's location. In the third row, the targeted person is walking at a different speed away from the camera wearer, and the camera wearer is also changing the walking path. In this case, only LIP-LSTM is able to reliably predict the targeted person's location at $t_{0}+10$. We hypothesize that it is because LIP-LSTM receives additional IMU information to make correct prediction. In the fourth row, the targeted person is walking across the camera wearer. In this case, although LR, L-LSTM, and LIP-LSTM correctly predict the person is passing from left to right, LIP-LSTM produces a more accurate final location at $t_{0}+10$. In the fifth row, the targeted person is standing still in the clip and the camera wearer is turning left as well as moving forward. Both STATS and LR fail to predict the person's future location and trajectory. While L-LSTM is able to predict the relative movements of the person in the future frames, it still fails to predict the person's final location at $t_{0}+10$. In this case, only the predictions of LIP-LSTM are close to the ground truth at both $t_{0}+5$ and $t_{0}+10$. Two failure cases are shown in the sixth and seventh rows. In the sixth row, although LIP-LSTM is able to predict the targeted person is walking toward the left, it fails to catch the walking speed of the targeted person, and therefore the predicted location and trajectory lag behind the ground truth. In the seventh row, the targeted person is standing still in the scene. Although the camera wearer is moving toward the targeted person, we hypothesize that the LIP-LSTM overestimates the walking speed of the camera wearer, and therefore fails to predict the future location of the targeted person at both $t_{0}+5$ and $t_{0}+10$ in this case.

\subsection{Results on Each Walking Direction}

As shown in the Table~\ref{tab:dataset}, the samples of the targeted person walking away from the camera wearer account for 80\% of the entire dataset, which may induce bias in network learning. Therefore, we further divide the dataset into 4 splits based on the walking direction. Each split has only one type of walking direction, and both L-LSTM and LIP-LSTM are retrained on each split. We use the same setting as in Section~\ref{subsec:implementation_details} to retrain them on the \textit{away} split. For the rest three splits, we use a batch size of 32, and train the networks for 1,000 epochs. Table~\ref{tab:res_each_walking_direction} shows the results on each split of different walking directions. In all scenarios, LIP-LSTM has the highest Mean IOU and lowest Mean DE. In the \textit{away} split, although the measured Mean IOU and Mean Final IOU of LR, L-LSTM, and LIP-LSTM are close, in terms of Mean DE, LIP-LSTM has a 3.4\% relative improvement with regard to L-LSTM, and 8.8\% relative improvement with regard to LR. It is also worth noting that all methods perform the best on the \textit{away} split compared to their performance on the other three splits. This may suggest that increasing the number of samples of the other three walking directions may lead to better performance on them, as the number of samples in the \textit{away} split is roughly 10 times more than the other three. On both the \textit{toward} and \textit{across} splits, although L-LSTM achieves the best Mean Final IOU, LIP-LSTM is able to reduce the Mean DE on both splits compared to L-LSTM, 6.6\% and 16.8\% relative improvements on the \textit{toward} and \textit{across} splits, respectively. On the \textit{still} split, LIP-LSTM outperforms the other three methods across all metrics, a 48.6\% relative increase compared to L-LSTM in the measured Mean Final IOU.

\begin{figure}[!t]
\centerline{\includegraphics[width=\linewidth]{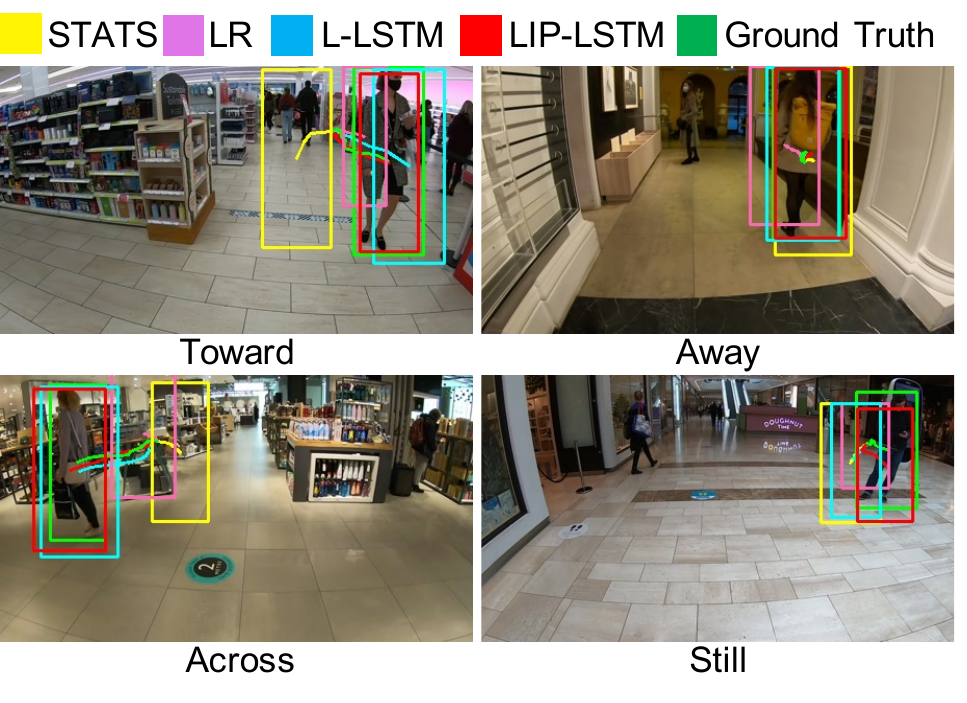}}
\caption{Qualitative results on each walking direction split. L-LSTM and LIP-LSTM are retrained on each split. We show one example for each split. The predicted future location (at $t_{0}+10$) and walking trajectory of a targeted person are rendered on the $t_{0}+10$ frame.}
\label{fig:vis_train_on_sep}
\end{figure}

Figure~\ref{fig:vis_train_on_sep} shows some qualitative results on each walking direction split. In the example of the targeted person walking toward the camera wearer, the predicted future location and trajectory of LIP-LSTM well match the ground truth. In the example of the targeted person walking away, both L-LSTM and LIP-LSTM make better predictions than STATS and LR. In the example at the bottom-left of Figure~\ref{fig:vis_train_on_sep}, the targeted person is walking across the camera wearer. Although the predicted future location of LIP-LSTM does not completely align with the ground truth, its predicted movement trajectory well matches the true trajectory. In the example of the targeted person staying still, LIP-LSTM also makes better predictions than the other three methods.

\section{Conclusions}

In this work, we address the task of predicting future person location and trajectory in egocentric videos captured by a wearable camera. An LSTM-based encoder-decoder framework is designed, which utilizes the cues of the past locations and body poses of the targeted person, and the past motion of the camera wearer recorded by the camera IMU. A new egocentric indoor dataset is constructed to evaluate the performance of our method. Both quantitative and qualitative results have shown that our method is able to reliably predict a targeted person's future location and trajectory relative to the camera wearer's viewpoint. The constructed new dataset can also be used to develop and evaluate the performance of a tracking system in the egocentric setting.

\bibliographystyle{IEEEtran}

\end{document}